# A chatbot architecture for promoting youth resilience

CHESTER HOLT-QUICK[a,b], JIM WARREN[a] [1], KAROLINA STASIAK[b], RUTH WILLIAMS[b], GRANT CHRISTIE[b], SARAH HETRICK[b], SARAH HOPKINS[b], TANIA CARGO[b], SALLY MERRY[b]
[a]*School of Computer Science* [b]*Department of Psychological Medicine, University of Auckland, Auckland 1142, NEW ZEALAND*

**Abstract.** E-health technologies have the potential to provide scalable and accessible interventions for youth mental health. As part of a developing an ecosystem of e-screening and e-therapy tools for New Zealand young people, a dialog agent, Headstrong, has been designed to promote resilience with methods grounded in cognitive behavioral therapy and positive psychology. This paper describes the architecture underlying the chatbot. The architecture supports a range of over 20 activities delivered in a 4-week program by relatable personas. The architecture provides a visual authoring interface to its content management system. In addition to supporting the original adolescent resilience chatbot, the architecture has been reused to create a 3-week 'stress-detox' intervention for undergraduates, and subsequently for a chatbot to support young people with the impacts of the COVID-19 pandemic, with all three systems having been used in field trials. The Headstrong architecture illustrates the feasibility of creating a domain-focused authoring environment in the context of e-therapy that supports non-technical expert input and rapid deployment.

## 1. Introduction

Most young people growing up in New Zealand report good overall wellbeing; however emotional difficulties are still prevalent, with young people reporting high levels of depressive symptoms, self-harming behavior and suicidal ideation [1]. Most mental health disorders have an onset during adolescence, a time associated with poor help-seeking and access to services [2] setting the scene for longer-term negative outcomes such as adverse mental health and economic outcomes in early adulthood [3]. Schools are aware of students' mental health issues and the lack of readily available interventions; 62% of principals reported being frustrated they could not get help for students with mental health issues and students in lower socio-economic areas were in greater need of support [4].

Providing e-Health interventions has shown to lead to improved emotional well-being in general [5] and in school settings [6]. E-Health interventions can provide near-universal reach and access, and potentially appeal to young people concerned about the stigma of using the conventional mental health system. Dialog agent or 'chatbot' style interaction for digital mental health has attracted interest since Eliza in the 1960's and evidence of effectiveness has been demonstrated for the modern system, Woebot [7].

---
[1] Corresponding Author

The Health Advances through Behavioural Intervention Technologies (HABITs) project is developing an ecosystem of screening and e-therapy tools designed to meet the needs of New Zealand young people with a co-design approach emphasizing input from Māori and Pacific youth. In keeping with this approach, the objective is to deliver evidence-based therapy in a form tailored to resonate with local young people. As part of HABITs we developed a chatbot providing dialog-based intervention grounded in cognitive behavioral therapy and positive psychology, called Headstrong. In this paper we present the architecture underlying the Headstrong chatbot, and related chatbots, starting with the conceptual design.

## 2. Conceptual Design of a Chatbot for Resilient Youth

A tender process was used to select a contractor with experience in innovative design and health IT. RUSH Digital were selected to partner with the university-based team and our community stakeholders to create the chatbot. A hybrid user-centered design / co-design process was used, beginning with scoping interviews of young people and experienced counselors. This was followed by design workshops including young people, counselors, researchers and developers. The resulting design focuses on daily user engagement in key activities over a 4-week intervention program, including: relaxation strategies; problem solving techniques; recognizing and tackling negative thoughts; interpersonal and communication skills, using a gratitude journal; and scheduling positive activities.

While Headstrong is fundamentally text dialog based, avatars were designed (see figure 1) to make the chat agent relatable. The avatars are rendered as young people slightly older and more experienced than the intended users, and giving a choice of gender and ethnicity representative of the target population. These avatars appear in various poses during the chat, such as providing a 'selfie' when initially building rapport with the user and with various expressions as appropriate to the messages in the activities.

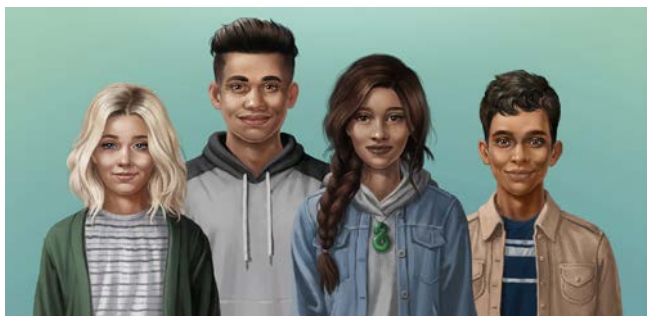

**Figure 1.** The four Headstrong avatars - from left: Olivia, Manaia, Aroha and Ravi.

## 3. Headstrong Architecture

Google's Dialogflow (https://dialogflow.com/) and other similar technologies such as Watson Assistant by IBM (https://www.ibm.com/cloud/watson-assistant/) offer simple off-the-shelf solutions to configure and deploy a chatbot and are becoming widely used. Such solutions are well suited to limited question-answering type help chatbots often employed in businesses, but configuring and managing (and visualizing) large amounts of content quickly becomes untenable. Implementing the Headstrong design required dialog encompassing more than 20 distinct activities each of which engages the user for 5-10 minutes. Such a rule-based dialog graph contains thousands of dialog nodes. We needed an architecture that facilitates managing this amount of content. Further, to streamline the knowledge engineering process, and to provide ease of adaptation for differing audiences and uses, authoring of dialog content should be possible by users without technical expertise. We wanted enhanced functionality within the conversational channel such as media or minigames; and for research requirements, we needed detailed but ethical tracking of chatbot usage.

Figure 2 illustrates the Headstrong architecture. Facebook Messenger was selected as the front-end so the chatbot could engage young people with a familiar interface likely to be already installed on their phones. The Headstrong server leverages existing components RUSH Digital had deployed to other industries, and uses Dialogflow for matching user input to intents. It interfaces to the HABITs digital platform [8] to record usage data for research. The browser-based interface for dialog authoring is described in the next section.

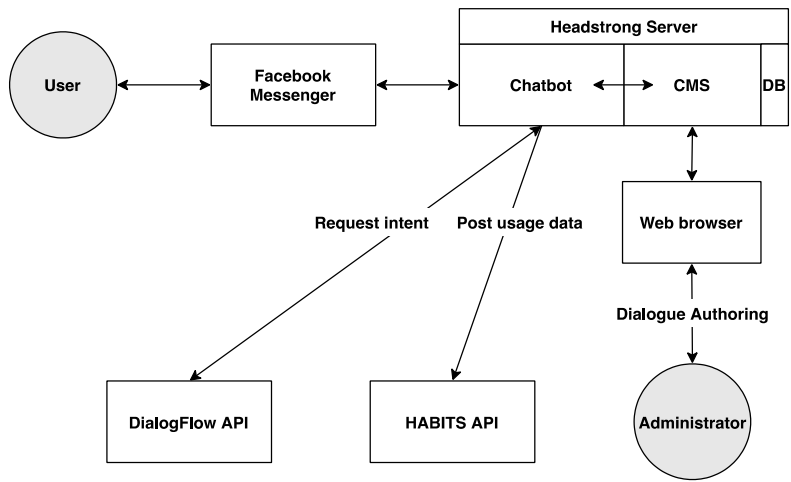

**Figure 2.** Components of the Headstrong architecture.

## 4. Content management and authoring

To give domain experts the ability to directly author content, a graphical user interface is provided that creates a directed graph structure. Node types are color-coded and dragged down from a toolbar into a working canvas. For example, Figure 3 includes a condition-checking node (pink), with conditions along each branch (beige), a question node (green) with quick reply options (beige) and two modules (lime). The '+' icon on the lower-right of some nodes can be clicked to add conditions or options to the parent node. The resulting graph is a logical flow diagram, intuitive to non-technical users.

There remain aspects of dialog authoring that are too challenging for non-technical users. One such example is configuring an appropriate chatbot introduction based on whether the chatbot is responding to unprompted user engagement, or handling a response from chatbot prompted engagement; and at the same time making appropriate reference to any important prior user state, such as whether the user had expressed distress and triggered the emergency escalation module during the prior engagement. This use case requires defining and managing user variables and setting up conditional branches to a degree that is generally just beyond the purview of a non-technical user.

Psychologists and other mental health specialists principally authored the activity modules. These modules consist predominantly of statement nodes (where text is output), questions with quick replies and output media files such as images and animated GIFs.

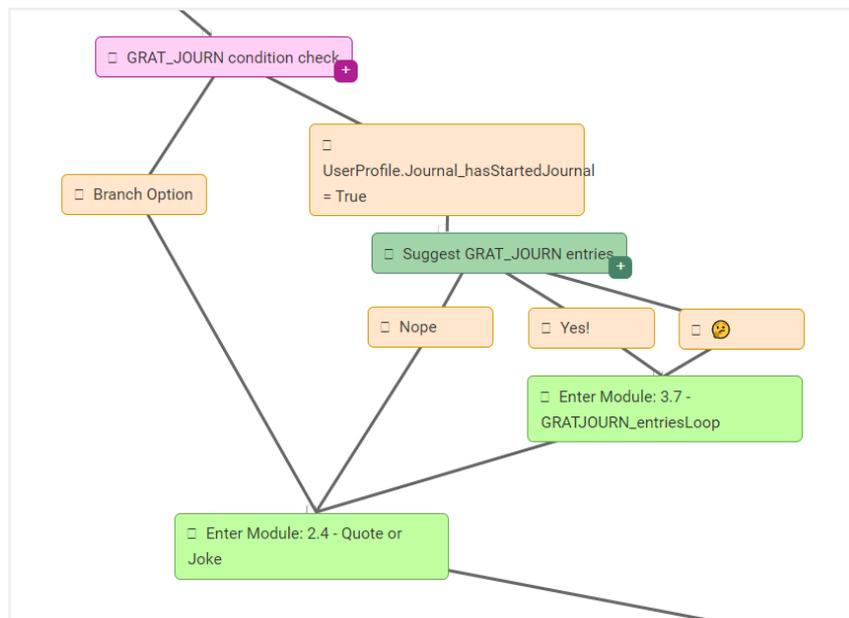

**Figure 3**. A portion of the daily loop routine as shown on the working canvas.

### 5. Use and Re-use of the Dialog Architecture

Figure 4 illustrates a snippet of a dialog session as it appears to the end user. After a period of iterative refinement based on feedback from both experts and young people, a field trial was set up using online consent in the HABITs portal. In late 2019, users were recruited from two local secondary schools. While the trial results are outside of the scope of this paper, it can be said that there were no issues from a technical perspective.

Here two technical features warrant mention. Firstly, the Headstrong architecture contains a dialog duplication function which allows the authored content of one dialog version to be copied, altered and re-used for deployment as another chatbot. Secondly, the Headstrong architecture supports the deployment of multiple chatbots with distinct dialog versions, connected to distinct channels (i.e. different Facebook pages) from one running server instance.

Even before the Headstrong field trial commenced, the architecture was re-used for another chatbot called "21-Day Stress Detox", developed as a Masters research project [9] with the intention of fostering stress resilience in undergraduate university students. The dialog content, spanning 21 days, was designed for users to engage with the chatbot once per day, wherein each day the chatbot would check in with the user at a chosen time (set at onboarding), ask them to rate their stress level, and then begin a new activity module. Each day ends with a choice of a motivational quote, joke, or an entry into a gratitude journal. The activity modules are unique for each day while other components of the 'daily loop' are re-used; although for variation chatbot statements such as introductions are drawn randomly from a pre-authored set. The chatbot was successfully field tested with over 120 undergraduates recruited from several large classes.

A moderate amount of dialog content from Headstrong was able to be reused for "21-Day Stress Detox" with minor modification tailored to a slight difference in target age group. Specifically, modules for greeting and farewell, minigames, jokes, meditation, gratitude journaling and risk phrase management were all reused; as was the intent classifier at the Dialogflow endpoint. Reused content constituted roughly one half of "21-Day Stress Detox".

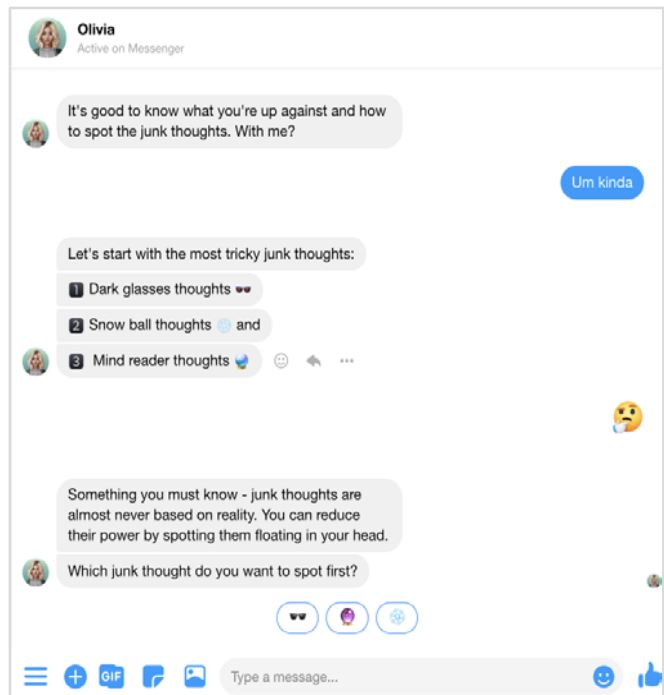

**Figure 4.** Screen capture of a fragment of Headstrong conversation within one of six cognitive modules targeted at negative thinking.

The novel dialog content was almost completely authored by the psychology Masters student in under 2 months. This required an initial 2-hour teaching session by a software developer, followed by two further sessions to build on the first session and troubleshoot any issues. Additionally, the developer set up the overall logical architecture for the program including more complicated aspects of authoring such as management of user variables and resolving errors. In total approximately 30 hours of developer time were required. Minimal developer time was required for authoring the unique daily activities.

This set the scene for the very rapid development and deployment of Aroha, a chatbot targeted to young people to support them with the various impacts related to the COVID-19 pandemic. The overall conversation design was newly created, tailored to support expected one-off user-driven engagement with Aroha, hence differing from the conversation design for Headstrong and 21-Day Stress Detox. Some content (i.e. gratitude journaling, minigames, meditation) from those two projects was re-used through module duplication, with minor context-sensitive modification to the dialog. The main new content for Aroha was developed by two experienced clinical psychologists over one weekend before lockdown in New Zealand. Dialog refinement, testing and implementation of feedback took place over 10 days and then a clinical trial begun.

**Discussion and Conclusion**

We have developed an architecture that supports easy authoring and deployment encompassing key requirements of a chatbot to promote resilience in young people, and have reported its use for three distinct deployments with different content and audiences. Through the graphical authoring canvas, content creation was largely accomplished by domain experts with technical support required only around areas such as definition of working variables to support branching and reminder logic.

Headstrong dialog is the result of expert adaptation of evidence-based psychological therapies and co-design with target users. In contrast, the performance of XiaoIce, a social chatbot emphasizing emotional connection that has communicated with over 660 million active users [10], shows that deep learning can be successfully applied to dialog. An outstanding challenge is integrating expert-crafted dialog with the engagement capability of deep-learning based systems in a single e-therapy agent. We believe there are possibilities in the medium-term including machine learning of dialog policy (e.g. sequencing activities) and use of deep-learnt dialog for rapport building and discovery while retaining crafted dialog for evidence-based activities. Integrating such deep learning-based capabilities is a key direction for development of the architecture with the objective of achieving longer retention in evidence-based programs.

In conclusion, the Headstrong architecture illustrates the feasibility of creating a domain-focused authoring environment for e-therapy. The architecture supports expert input through its visual interface to the content management system. Further, the architecture allows rapid deployment to field studies and sufficient flexibility to support interventions of different lengths and for different target audiences. Work continues to enhance the chatbot's ability to tailor responses for maximum user engagement.

**Acknowledgements**

We thank all the participants in the development process and field trials. The authors have no commercial interest in RUSH Digital but may be designated inventors in Intellectual Property relating to novel dialog components and/or content described in this paper. This work was funded by A Better Start (grant UOAX1511) and CureKids (Discretionary grant 5050).

## References

[1] T. Clark *et al.*, "Health and well-being of secondary school students in New Zealand: Trends between 2001, 2007 and 2012," *Journal of Paediatrics and Child Health,* vol. 49, no. 11, pp. 925-934, 2013.
[2] N. Reavley, S. Cvetkovski, A. Jorm, and D. Lubman, "Help-seeking for substance use, anxiety and affective disorders among young people: results from the 2007 Australasian National Survey of Mental Health and Wellbeing," *Australian & New Zealand Journal of Psychiatry,* vol. 44, no. 8, pp. 729-735, 2010.
[3] D. Fergusson, J. M. Boden, and L. J. Horwood, "Recurrence of major depression in adolescence and early adulthood, and later mental health, educational, and economic outcomes," *British Journal of Psychiatry,* vol. 191, pp. 335-342, 2007.
[4] L. Bonne and J. MacDonald, "Secondary Schools in 2018: Findings from teh NZCER National SUrvey," New Zealand Council for Educational Research, Wellington2019.
[5] D. D. Ebert *et al.*, "Internet and computer-based cognitive behavioral therapy for anxiety and depression in youth: a meta-analysis of randomized controlled outcome trials.," *PLoS ONE* vol. 10, no. 3, p. e0119895., 2015.
[6] Y. Perry *et al.*, "Trial for the Prevention of Depression (TriPoD) in final-year secondary students: study protocol for a cluster randomised controlled trial," *Trials [Electronic Resource],* Research Support, Non-U.S. Gov't vol. 16, p. 451, 2015.
[7] K. K. Fitzpatrick, A. Darcy, and M. Vierhile, "Delivering Cognitive Behavior Therapy to Young Adults With Symptoms of Depression and Anxiety Using a Fully Automated Conversational Agent (Woebot): A Randomized Controlled Trial," *JMIR Ment Health,* vol. 4, no. 2, p. e19, Jun 6 2017.
[8] J. Warren, S. Hopkins, A. Leung, S. Hetrick, and S. Merry, "Building a digital platform for behavioural internvention technology research and deployment," in *Proc 53rd Hawaii International Conference on System Sciences*, 2020.
[9] R. Williams, "Development of a pilot trial of a chatbot as a digital wellbeing intervention to reduce stress in tertiary students," Thesis submitted for Master of Science in Psychology, University of Auckland, 2020.
[10] L. Zhou, J. Gao, D. Li, and H.-Y. Shum, "The Design and Implementation of XiaoIce, an Empathetic Social Chatbot," *Computational Linguistics,* pp. 1-62, 2020.